\documentclass{article}

\usepackage[final, dblblindworkshop]{neurips_2025}

\usepackage[utf8]{inputenc} 
\usepackage[T1]{fontenc}    
\usepackage{hyperref}       
\usepackage{url}            

\usepackage{booktabs}       
\usepackage{adjustbox}
\usepackage{amsfonts}       
\usepackage{amsmath}
\usepackage{nicefrac}       
\usepackage{microtype}      
\usepackage{xcolor}         
\usepackage{graphicx}
\usepackage{multirow}
\usepackage{comment}
\usepackage{graphicx}
\usepackage{subcaption}
\usepackage{booktabs}
\usepackage{float}
\usepackage{placeins}
\usepackage{afterpage}
\usepackage{needspace}

\title{Realistic CDSS Drug Dosing with End-to-end Recurrent Q-learning for Dual Vasopressor Control} 
\workshoptitle{NeurIPS 2025 Workshop on Time Series for Health}

\author{%
 \textbf{Will Y. Zou}$^{2}$ \quad 
  \textbf{Jean Feng}$^{1}$ \quad
  \textbf{Alexandre Kalimouttou}$^{1}$ \\
  \textbf{Jennifer Yuntong Zhang}$^{2,3}$ \quad
  \textbf{Christopher W. Seymour}$^{4}$ \quad
  \textbf{Romain Pirracchio}$^{1}$ \\
  $^{1}$University of California, San Francisco \quad 
  $^{2}$Angle.ac \\
  $^{3}$Engineering Science, University of Toronto \quad
  $^{4}$University of Pittsburgh\\[0.2cm]
}

\begin{document}

\maketitle
\vspace{-0.4cm}
\begin{abstract}
\vspace{-0.2cm}
Reinforcement learning (RL) applications in Clinical Decision Support Systems (CDSS) frequently encounter skepticism because models may recommend inoperable dosing decisions.
We propose an end-to-end offline RL framework for dual vasopressor administration in Intensive Care Units (ICUs) that directly addresses this challenge through principled action space design.
Our method integrates discrete, continuous, and directional dosing strategies with conservative Q-learning and incorporates a novel recurrent modeling using a replay buffer to capture temporal dependencies in ICU time-series data. Our comparative analysis of norepinephrine dosing strategies across different action space formulations reveals that the designed action spaces improve interpretability and facilitate clinical adoption while preserving efficacy. Empirical results on \emph{eICU} and \emph{MIMIC} demonstrate that action space design profoundly influences learned behavioral policies. Compared with baselines, the proposed methods achieve more than $3\times$ expected reward improvements, while aligning with established clinical protocols \footnote{Code is publicly available at \text{https://github.com/wzoustanford/vaso\_rl/}}. 
\vspace{-0.2cm}
\end{abstract}

\vspace{-0.3cm}
\section{Introduction}
\vspace{-0.2cm}


Clinical interventions in ICUs are inherently heterogeneous, spanning discrete choices and continuous adjustments~\cite{jayaraman2024primer,eghbali2021patient,qiu2022latent,tosca2024model}. 
This diversity places strong demands on action space in Reinforcement Learning (RL): they should capture the full spectrum of clinical practice while remaining tractable for learning. 
Without careful action space design, RL policies risk generating unrealistic or impractical recommendations that clinicians are unlikely to adopt. 
Specifically, in  Clinical Decision Support System (CDSS)~\cite{hunt1998effects,garg2005effects} for septic shock, clinicians apply norepinephrine (first-line vasopressor) and vasopressin (second-line vasopressor) over multiple phases—initiation, titration, and weaning.
While reinforcement learning offers a natural framework for this sequential decision-making problem, prior work~\cite{kalimouttou2025optimal,komorowski2018artificial} often focuses on single vasopressor control and early treatment phases, with fewer examining how titration evolves throughout the entire treatment period.


In this work, we study the effectiveness of action space design in influencing the learned dosing policies. 
Concretely, we adopt the offline settings of the Deep Q-learning (DQN)~\cite{mnih2015human} with recurrent replay~\cite{kapturowski2018recurrent} to capture temporal dependencies, and incorporate conservative Q-learning (CQL)~\cite{kumar2020conservative} to ensure policy safety. Integrating them together as a potential CDSS system, we validate the action space design for dosing in treatment trajectories. Using this integrated offline RL framework as a CDSS, we evaluate discrete, continuous, and directional dosing action spaces and analyze their impact on \emph{both norepinephrine and vasopressin} control over the \emph{full treatment horizon}.


Our results show that discrete and directional-discrete action spaces are not only more interpretable, aligned with clinician expectations, but they also significantly improve model performance compared to continuous dosing. Their sparsity and higher dimensionality, combined with the recurrent replay framework enable more generalizable models that achieve more than $3\times$ expected reward improvements with weighted importance sampling offline-policy evaluation. These results show the potential of deep reinforcement learning and action space design for building effective CDSS systems.

\vspace{-0.3cm}

\section{Prior Work} 
\vspace{-0.2cm}
Reinforcement learning techniques have been applied in several independent studies for treating sepsis shock. For instance,~\cite{saria2018individualized} proposed individualized treatment for sepsis using RL, and \cite{komorowski2018artificial} provided an earlier comprehensive study of sepsis treatment with publicly available datasets. More recently,~\cite{jayaraman2024primer} studied ICU data and RL in medicine. \cite{kalimouttou2025optimal} provided a focused study of effective dosing action space design in an end-to-end RL system for ICU sepsis treatment. 

For action space design in RL, earlier work addressed factored~\cite{tavakoli2018action}, hierarchical~\cite{kulkarni2016hierarchical}, continuous action spaces~\cite{tessler2019action,delalleau2019discrete}.~\cite{kapturowski2018recurrent} addressed the importance of recurrent sequence learning in replays for application in Atari games, and ~\cite{schrittwieser2020mastering} progressed recurrent learning into model-based RL applications. Offline RL methods such as conservative Q-learning~\cite{kumar2020conservative}, implicit Q-learning \cite{kostrikov2021offline} and advantage weighted Actor-Critic \cite{nair2020awac} applied different regularization and policy constraints. Model-based offline RL approaches \cite{kidambi2020morel} focused on modeling transition dynamics. In the clinical literature, \cite{eghbali2021patient} applied RL algorithm to sedation dosing in ICUs. \cite{tosca2024model} provided acollection of dosing techniques focusing on referencing cancer and chemotherapy treatments. 
~\cite{qiu2022latent} applied a latent constrained batch algorithm to address the risks of incorrect dosing heparin in ICU. However, the use of RL to systematically design and evaluate vasopressor dosing strategies for sepsis shock remains under-explored. 


\vspace{-0.3cm}
\section{Algorithms}
\label{sec3:algorithms}
\vspace{-0.2cm}
We propose an offline Q-learning framework that employs dosing action space design with a conservative Q-learning objective, and incorporates a recurrent replay buffer to model ICU time-series.

\vspace{-0.3cm}
\subsection{Dosing Action Space Design} 
\label{sec: action_design}
\vspace{-0.2cm}

We introduce an action space design that bridges the gap between reinforcement learning optimization and clinical feasibility. The action space contains a binary decision for vasopressin ($vp_1$) and dosing decisions for norepinephrine ($vp_2$) which takes values in the range $(0, 0.5]$ (mcg/kg/min)\footnote{Due to international clinical practice, we ensure the dosing is above zero for the first-line vasopressor.}. 
Specifically, three action space designs for $vp_2$ are proposed: continuous dosage, clinically-aligned discrete intervals based on standard protocols, and directional dose adjustments that mirror clinical decision-making. Our action space $\mathcal{A}$=\{($vp_1$,$vp_2$)\} models dual vasopressor administration, maintaining model performance while preserving the tractability of real-world intensive care interventions. This design enables deployment within existing clinical workflows while maintaining the expressiveness required for effective policy learning.

\emph{\textbf{Continuous dosing}}. A continuous dosing variable is applied for vasopressor norepinephrine allowing it to freely move in a range. The action space is $\mathcal{A} =\{(vp_{1}, vp_{2})\}$, where $vp_{1}\in\{0, 1\}$ is a binary variable, and $vp_{2}\in(0, 0.5]$ (mcg/kg/min).

\emph{\textbf{Block discrete dosing}}. We define the dosing action to be the discrete values in the same range as in the continuous dosing. The discrete values only vary across meaningful discrete dosing blocks. We explore different numbers of blocks to validate the impact of effective dosing. The action space is $\mathcal{A} =\{(vp_{1}, vp_{2})\}$, where $vp_{2}\in \text{range}(\delta/2, 0.5, \delta)$ (mcg/kg/min) and $\delta$ is bin size computed from the number of bins.

\emph{\textbf{Stepwise directional dosing}}. We define the the action space at each time step to be a directional and stepwise change. The action space is $\mathcal{A} =\{(vp_{1}, vp_{2})\}$, where $vp_{2}$ is a stepwise change from the current dose. For example, $vp_{2}\in \{-0.1, -0.05, 0, +0.05, +0.1\}$ (mcg/kg/min). The state space is extended to include a one-hot vector that stores and keeps track of the current discrete dosage of norepinephrine.  

\vspace{-0.3cm}
\subsection{Q-Learning for Vasopressor Control}
\vspace{-0.2cm}
We formulate the vasopressor control problem with $\mathcal{M} = (\mathcal{S}, \mathcal{A}, \mathcal{P}, \mathcal{R}, \gamma)$, where $\mathcal{S}$ is the state space, $\mathcal{A}$ is the action space of vasopressor decisions, $\mathcal{P}: \mathcal{S} \times \mathcal{A} \times \mathcal{S} \rightarrow [0,1]$ is the state transition dynamics, $\mathcal{R}: \mathcal{S} \times \mathcal{A} \rightarrow \mathbb{R}$ is the reward, and $\gamma \in [0,1]$ is the discount factor. The state $s_t \in \mathcal{S} \subseteq \mathbb{R}^d$ comprises patient information, physiological measurements \footnote{The measurements include mean blood pressure (MBP), lactate levels, blood urea nitrogen, serum creatinine, (total) fluid, urine of the hour, corticosteroid, and Sequential Organ Failure Assessment (SOFA) score.} , and organ support \footnote{The support includes mechanical ventilation and renal replacement therapy.}. The action space $\mathcal{A}$ contains a binary decision for vasopressin ($vp_1$) and dosing decisions for norepinephrine ($vp_2$). The dosing decisions are formulated in Section \ref{sec: action_design}. The reward function contains two objectives: maximizing the chance of survival and keeping vasopressors in dosage limit. Given a dataset $\mathcal{D} = \{(s_t, a_t, r_t, s_{t+1})\}$ of historical ICU trajectories collected under behavioral policy $\pi_b$ (clinical practice), our goal is to learn an optimal policy $\pi^*$ that maximizes expected cumulative reward. 

\textbf{Q-Learning.} Q-learning seeks to learn the action-value function $Q(s,a)$ representing the expected cumulative reward by taking action $a$ in state $s$ and following the optimal policy thereafter. The Q-function satisfies the Bellman optimality: $Q^*(s,a) = \mathcal{R}(s,a) + \gamma \mathbb{E}_{s' \sim \mathcal{P}(\cdot|s,a)}[\max_{a'} Q^*(s',a')]$. In the offline setting, Q-learning minimizes the temporal difference (TD) mean-squared error:
\vspace{-0.1cm}
\begin{align}
\label{eq:td_obj}
\mathcal{L}_{\text{TD}}(\theta) = \mathbb{E}_{(s,a,r,s') \sim \mathcal{D}} \left[ \left( Q_\theta(s,a) - (r + \gamma \max_{a'} Q_{\theta^-}(s',a')) \right)^2 \right],
\end{align}
where $Q_\theta$ is the learned Q-network with parameters $\theta$, and $Q_{\theta^-}$ is a target network updated periodically for stability. We apply a version of double Q-learning~\cite{van2016deep} with two Q networks and two corresponding target networks. 

As an improvement to Q-learning, we optionally add a conservative  term~\cite{kumar2020conservative} (conservative Q-Learning) to penalize Q-values for out-of-distribution actions, with $\alpha$ as strength of the penalty:
\begin{align*}
\mathcal{L}_{\text{CQL}}(\theta) = \alpha \cdot \mathbb{E}_{s \sim \mathcal{D}} \left[ \log \sum_{a} \exp(Q_\theta(s,a)) - \mathbb{E}_{a \sim \mathcal{D}(a|s)}[Q_\theta(s,a)] \right] + \mathcal{L}_{\text{TD}}(\theta).
\end{align*}

\vspace{-0.3cm}
\subsection{Recurrent Experience Replay with Effective Dosing} 
\vspace{-0.2cm}
Our approach applies the recurrent experience replay algorithm~\cite{kapturowski2018recurrent} to address the temporal dependencies in ICU vasopressor control. The algorithm employs a prioritized sequential buffer $\mathcal{RB}=\{\mathcal{S}_e\}$, storing  treatment episodes $\mathcal{S}_e=\{(s_t, a_t, r_t, s_{t+1})_{t\in e}\}$, to capture time-sensitive patterns of patient responses to vasopressor adjustments. In contrast to the sample-wise Q-function application in traditional Q-learning, we leverage an LSTM network to model the sequential nature of ICU states and interventions, recognizing that vasopressor effects unfold over time. In addition, we incorporate the \emph{sum-tree} data structure to prioritize episodes with larger temporal difference (TD) errors~\cite{schaul2015prioritized}, improving learning in challenging clinical scenarios. The recurrent model combines with specialized action space design to improve reinforcement learning for patient stability.


\vspace{-0.3cm}
\section{Experiments}
\vspace{-0.2cm}
We conduct multiple experiments demonstrating that our action space designs are both learnable and effective. We first validate learnability and model behavior through pilot evaluations with $\Delta$ Q-value and Fitted Q-Evaluation, using a simplified reward (Section \ref{subsec:pilot_exp}). We then show concrete gains through full off-policy evaluation under a comprehensive clinical reward (Section \ref{subsec:main_exp}).

\vspace{-0.3cm}
\subsection{Learnability of Clinically Aligned Action Spaces}
\label{subsec:pilot_exp}
\vspace{-0.2cm}
We evaluate our dosing action space design with the reward function defined in Table~\ref{tab:simple_reward_definition}. The objective is to evaluate whether each action space formulation supports meaningful value learning by tracking Q-value improvements and estimating policy quality using Fitted Q-Evaluation (FQE) on sepsis treatment data from eICU~\cite{pollard2018eicu} and MIMIC-IV~\cite{johnson2023mimic} datasets \footnote{See Appendix \ref{apdx:data} for more data and processing details.}. 

\begin{table}[ht!]
\centering
\begin{tabular}{lc}
\toprule
Type & Rule of the reward function \\
\midrule
Mortality: & $+10$ if the patient survived, and $-10$ otherwise\\
Vaso. penalty: &  At each time step, aggregate to reward: $-0.1 \cdot (vp_1 + vp_2 \cdot 2 - 1.0))$\\
\bottomrule
\end{tabular}
\vspace{0.2cm}
\caption{Definition of the simple reward function.}
\vspace{-0.3cm}
\label{tab:simple_reward_definition}
\end{table}


\textbf{Action Spaces.} Following Section \ref{sec: action_design}, we compare five formulation of action space: \textbf{(i)} the Binary $vp_1$ model, which treats vasopressin as a binary variable; \textbf{(ii)} the Dual Mixed model, which presents $vp_2$ as a continuous variable in the range $(0,0.5]$ (mcg/kg/min); \textbf{(iii)} the Dual Block Discrete (Dual BD) model, which discretizes $vp_2$ into $3$, $5$, or $10$ bins; \textbf{(iv)} the Dual Stepwise model, which encodes directional dose adjustments $vp_{2s}\in [{-\text{max\_step}, ..., -0.05, 0, +0.05, ..., +\text{max\_step}}]$ (mcg/kg/min); \textbf{(v)} the LSTM Block Discrete (LSTM BD), which discretizes $vp_2$ into $10$ bins trained with sequence length $5$ and a $2$-step burn-in. 

\textbf{RL Model Implementation.} The RL model leverages a fully-connected network as a Q-function which takes the combined state-action as input. For double-Q learning, during training and inference, Q-value is taken as the minimum of the target outputs, $q=\text{min}(q_1, q_2)$. The Q networks are periodically updated by adding weighted parameters of the target networks. The model is trained with the TD error objective (Equation~\ref{eq:td_obj}) \footnote{See Appendix \ref{apdx_sec:model_imp_training} and \ref{apdx_sec:model_arch} for model implementation and architecture. See Appendix \ref{apdx_sec:hyperparameters} for hyper-parameters.}. 

\textbf{$\Delta$ Q Evaluation.} As shown in  Table~\ref{tab:delta_q_comparison}, 
action space designs are evaluated with Q improvement per transition ($\Delta \text{Q /step}$), where $\Delta Q$ is the gain of model-recommended actions over clinician actions. 
We also measure concordance (C.) between model recommendations and clinician actions. 
For $vp_1$, concordance is binary agreement; for $vp_2$, concordance holds when the recommended dosage falls within the defined bin edges.

\vspace{-0.4cm}
\setlength{\tabcolsep}{2.5pt} 
\begin{table}[ht]
\centering
\caption{\textbf{Comparison of RL models.} We include Binary $vp_1$, Dual Mixed, Dual BD, Dual Stepwise, and LSTM BD models with the best  conservatism levels ($\alpha$) selected with the validation set.}
\label{tab:delta_q_comparison}
\begin{tabular}{llcccccc}
\toprule
Model & Config & $\alpha$ & $vp_1$ (\%) & Q/step & $\Delta$\textbf{Q/step} & $vp_1$ C.(\%) & $vp_2$ C.(\%) \\
\midrule
Clinician & -- & -- & 38.8 & -- & 0.000 & 100.0 & -- \\
\midrule
Binary $vp_1$ & -- & 0.001 & 79.8 & 0.869 & 0.028 & 44.5 & -- \\
Dual Mixed & -- & 0.010 & 77.8 & 1.345 & 0.185 & 52.5 & -- \\
\midrule
Dual BD. & 3 bins & 0.010 & 76.8 & 1.576 & 0.127 & 53.1 & 56.3 \\
Dual BD. & 5 bins & 0.000 & 96.3 & 2.383 & 0.265 & 39.7 & 26.4 \\
Dual BD & 10 bins & 0.000 & 92.3 & 4.673 & \textbf{0.309} & 42.4 & 13.3 \\
\midrule
Dual Stepwise &max\_step 0.1& 0.0001 & 12.4 & 0.094 & 0.136 & 65.1 & 17.7 \\
Dual Stepwise &max\_step 0.2& 0.0000 & 97.3 & 4.021 & \textbf{0.511} & 38.1 & 11.7 \\
\midrule
LSTM BD & 10 bins & 0.0000  & 100.0 & 0.878 & \textbf{0.456} & -- & 24.5 \\
LSTM BD & 10 bins & 0.0001  & 100.0 & 0.866 & 0.398 & -- & 20.4 \\
\bottomrule
\end{tabular}
\end{table}
\textbf{Fitted Q-Evaluation \footnote{See Appendix \ref{apdx:fqe} and \ref{apdx_sec:fqe_figures} for detailed FQE descriptions and figures .}}. FQE is applied as an off-policy evaluation (OPE)~\cite{zhang2022off,uehara2022review} method for all models. For each model, we compute test-set Q-values and fit a Gaussian to summarize their distribution. We compare model optimal policy with the clinician actions. As shown in Table~\ref{tab:fqe_ope}, block discrete models perform as well as continuous variable for $vp_2$. Stepwise directional with a larger max\_step has the largest $25.1\%$ probability of improved rewards. Moreover, incorporating an LSTM with recurrent experience reply yields a $17.4\%$ probability of improved rewards, which shows the benefits of capturing cross-time dependencies. 
\vspace{-0.3cm}
\setlength{\tabcolsep}{2.5pt} 
\begin{table}[ht]
\centering
\caption{\textbf{FQE OPE Results.} PIR (Probability of Improved Rewards) is defined as the probability of model recommended action reaching a higher reward value than the clinician mean reward, minus $50\%$ ($P$(Model > Cli. Mean) - $50\%$); $\Delta$ Q Mean is the shift in the mean of fitted Gaussian for Model vs Clinician.}
\label{tab:fqe_ope}
\begin{tabular}{lccccccc}
\toprule
Model & $\alpha$ & Config & PIR & $\Delta$ Q Mean & Mod. Mean & Cli. Mean & Cohen's d \\
\midrule
Binary $vp_1$ & 1e-3 & -- & 0.8\% & 0.028 & 0.869 & 0.841 & 0.019 \\
Dual Mixed &1e-2 & -- & 6.3\% &0.186& 1.346 &1.161 &0.146 \\
Dual BD &1e-2 & 3 bins & 3.0\% &0.127 & 1.576 &1.450 &0.073 \\
Dual BD &0.0& 5 bins &4.4\%&0.265&2.383&2.119&0.106\\
Dual BD &0.0& 10 bins &6.4\% &0.309 &4.673 &4.364 &0.151 \\
Dual Stepwise &0.0& max\_step 0.2 &\textbf{25.1\%} &\textbf{0.511} &4.021 &3.510 &\textbf{0.503} \\
LSTM BD & 1e-4 & 10 bins & \textbf{17.4\%} & 0.398 & 0.866 & 0.468 &\textbf{0.439} \\
\bottomrule
\end{tabular}
\end{table}

\vspace{-0.3cm}
\subsection{Performance with Full Clinical Reward and Weighted Importance Sampling (WIS)}
\vspace{-0.2cm}
\label{subsec:main_exp}
To show that the action space design concretely improves model performance, we implement the full reward function in \cite{kalimouttou2025optimal}, and leverage WIS for Offline-Policy Evaluation (OPE). 

\textbf{Comprehensive Reward Function}. Following \cite{kalimouttou2025optimal}, we implement a comprehensive reward function with adjusted rewards that indicate patient progress. The reward is composed of a base survival reward, a set of benefit for improved clinical parameters (such as improved mean blood pressure or lactate levels), reward for decreased norepinephrine usage, and mortality penalty \footnote{See Appendix \ref{apdx:crf} for more detailed reward description.}. 

\textbf{Weighted Importance Sampling}. OPE is performed on the test-set trajectories using. First, we identify optimal actions recommended by our model and train a softmax model on both the target policy (optimal model actions, $\pi_m$) and baseline policy (clinician actions, $\pi_c$). Then, the trajectory-level WIS coefficients are computed as specified in the first line of Equation~\ref{eq:stable_wis} to estimate the trajectory-level expected reward $R^{(i)}_{\text{traj}}$. Lastly, the overall WIS expected reward is computed using the renormalization shown on the second line of Equation~\ref{eq:stable_wis}.
\begin{equation}
\label{eq:stable_wis}
\begin{aligned}
R^{(i)}_{\text{traj}} &= 
T \cdot \frac{\sum_{t=0}^T w_t R_t}{\sum_{t=0}^T w_t},
&& \text{where } 
w_t = \frac{\pi_m(a_t^{\text{cli.}} \mid s_t)}{\pi_c(a_t^{\text{cli.}} \mid s_t)}, \\
R_{\text{WIS}} &= 
\frac{\sum_{i=0}^N \tilde{w}_i R_{\text{traj}}^{(i)}}{\sum_{i=0}^N \tilde{w}_i},
&& \text{where }
\tilde{w}_i = \frac{1}{T}\sum_{t=0}^T w_t.
\end{aligned}
\end{equation}

We also calculate the trajectory-level WIS by computing the trajectory-level weight as the product of transition level weights across the trajectory time-steps. This alternative evaluation resulted in similar ranges of WIS improvements as in \cite{kalimouttou2025optimal}. 
For our clinically-aligned discrete action spaces, we found the two-step WIS method described in Equation~\ref{eq:stable_wis} to be reliable and stable. 


Model implementation, architecture, and hyperparameters are the same as Section \ref{subsec:pilot_exp} except that we remove conservative Q-learning ($\alpha=0.0$), and train all models for $500$ epochs.

\textbf{Results.} As shown in Table~\ref{tab:wis_ope}, finer discretization in the action space yields better performance, and adding recurrent LSTMs with experience replay results in further substantial improvements. Both the Dual BD and LSTM BD models show over $3\times$ improved WIS expected rewards, compared with the the Dual Mixed model where $vp_2$ is a continuous variable.

\setlength{\tabcolsep}{2.0pt} 
\vspace{-0.4cm}
\begin{table}[ht]
\centering
\caption{\textbf{Importance Sampling (IS) and Weighted Importance Sampling (WIS) OPE Results.} We show transition-level (Trans. Lvl) average reward $E(R)$, IS, WIS, and corrsponding trajectory-level (Traj. Lvl) average reward. We also present the improvements in trajectory-level WIS (Imp. WIS) between model recommended actions and clinician actions. Lastly, we show the $95\%$ Confidence Interval (CI) for WIS improvement obtained by bootstrapping across test-set trajectories.}
\label{tab:wis_ope}
\begin{tabular}{lcccccccccc}
\toprule
\multicolumn{1}{c}{Model} & \multicolumn{1}{c}{Config} & \multicolumn{3}{c}{Trans. Lvl} & \multicolumn{3}{c}{Traj. Lvl}& \multicolumn{1}{c}{Imp. WIS}&\multicolumn{1}{c}{Imp.$\times$}&\multicolumn{1}{c}{95\% CI}\\
 &  & $E[R]$ & IS  &WIS& $E[R]$ & IS & WIS & (Mod.-Cli.) &\\
\midrule
Binary $vp_1$ &-- & 2.13 & 2.10 &2.15&112.01 & 110.51 & 117.54 & 5.53 & - &[2.0, 9.1] \\
Dual Mixed & -- & 2.13 &1.71& 2.23 &112.01 &89.69 &118.85 &6.84 &1$\times$&[-0.8, 14.2] \\
Dual BD & 3 bins & 2.13 &1.92 & 2.22&112.01 & 101.13 &120.98 &8.98 & 1.3$\times$&[2.6, 15.3] \\
Dual BD & 5 bins &2.13 &2.23&2.38&112.01&117.25 &139.80 &27.79 & 4.1$\times$&[19.8, 35.8] \\
Dual BD & 10 bins &2.13 &2.34 &2.40&112.01 &123.48 &141.51 &29.50 & 4.3$\times$ &[20.1, 38.3] \\
LSTM BD & 10 bins & 2.13 & \textbf{2.51} & \textbf{2.73} &112.01 &\textbf{131.83} &\textbf{155.41} &\textbf{43.40} & \textbf{6.3}$\times$&[33.6, 54.1] \\
\bottomrule
\end{tabular}
\end{table}

\vspace{-0.3cm}
\section{Conclusion}
\vspace{-0.2cm}
We establish action space design as a critical bridge between reinforcement learning optimization and clinical assistance in ICU settings. By developing a novel action space, we convert theoretical optimal policies into actionable clinical protocols in our proposed dual vasopressor control system for septic shock. Combined with recurrent experience replay to capture the complex temporal dependencies, our action space design achieves more than 3$\times$ expected reward improvements with WIS offline-policy evaluation compared with baselines. These contributions move towards a blueprint for CDSS systems that clinicians could interpret and trust, and towards AI systems that not only compute optimal actions, but deliver them in forms that integrate into critical medical workflows.
\bibliographystyle{unsrt}
\bibliography{neurips_workshop_timeseries}

\clearpage
\appendix
\section*{Appendix}

\section{Data Source and Processing}
\label{apdx:data}
The patient features extracted from the eICU and MIMIC-IV databases serve as the state representation for the reinforcement learning (RL) algorithm, and the action space consists of vasopressin administration (binary) and norepinephrine dosing in the continuous range $(0,0.5]$ (mcg/kg/min). The data are preprocessed such that missing values are imputed using forward filling, and norepinephrine dosages are clipped to the valid clinical range $(0,0.5]$. We include unique patients experiencing their first episode of septic shock, defined using the Sepsis-3 criteria \cite{singer16} and who were already receiving norepinephrine at the moment shock onset was identified. Shock onset could occur either while the patient was still in the emergency department or after ICU admission.

\section{Fitted Q-Evaluation (FQE)}
\label{apdx:fqe}
To evaluate the learned policies against historical clinician decisions without online deployment, we employ Fitted Q-Evaluation (FQE) with Gaussian distribution modeling. This approach computes Q-values for both the learned policy and clinician actions on the held-out test set, enabling direct comparison of expected returns. For each trained model, we extract Q-values by evaluating state-action pairs from the test trajectories where actions are either from the learned policy (model Q-values) or from historical clinician decisions (clinician Q-values). The resulting Q-value distributions are then fitted with Gaussian distributions using maximum likelihood estimation, providing parametric representations characterized by mean ($\mu$) and standard deviation ($\sigma$) for both the model and clinician policies. This parametric approach enables computation of key metrics including the mean improvement gap (difference in expected returns), probability of improvement (likelihood that model Q-values exceed clinician Q-values), and effect size (Cohen's d) to quantify the magnitude of policy differences.

The FQE analysis also generates visualizations including overlapping histograms with fitted Gaussian curves, cumulative distribution functions (CDFs) showing the probability mass distribution of Q-values, and improvement probability curves that illustrate the likelihood of the model outperforming various Q-value thresholds. The probability of improvement at the clinician's mean Q-value serves as a critical metric, indicating how often the learned policy is expected to achieve better outcomes than historical treatment decisions. Additionally, the analysis computes Cohen's d as a standardized effect size measure. The larger the value of Cohen's d, the more meaningful are the difference between the means of two groups. This framework provides quantitative evidence for policy improvement while accounting for the inherent uncertainty in Q-value estimates. The FQE analysis was performed for policy performance across different model architectures and hyperparameter configurations. The FQE numerical results are given in Table~\ref{tab:fqe_ope}. The histogram and sigmoid CDF visualizations are shown in Appendix~\ref{apdx_sec:fqe_figures}. 

\FloatBarrier
\Needspace{0.9\textheight}  
\section{Fitted Q-Evaluation Illustrations}
\label{apdx_sec:fqe_figures}

\begin{figure}[h]
\caption{FQE comparison between learned patient model and clinician baseline (Binary $vp_1$).}
\centering
\includegraphics[width=0.70\textwidth]{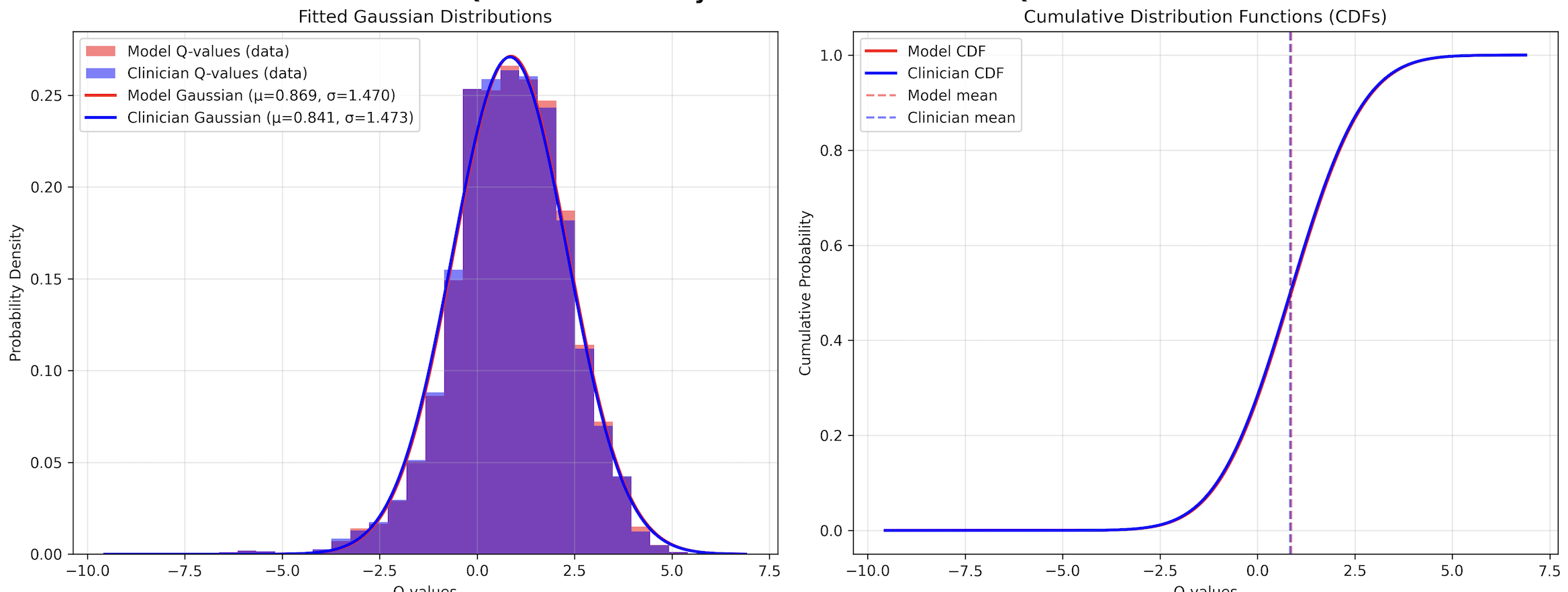}
\caption*{Binary $vp_1$ Model: distribution of Q-values from model (red) and clinician (blue).}
\label{fig:q_histogram_binary}
\end{figure}

\begin{figure}[h]
\caption{FQE comparison between the learned patient model and clinician baseline (Dual Mixed).}
\centering
\includegraphics[width=0.70\textwidth]{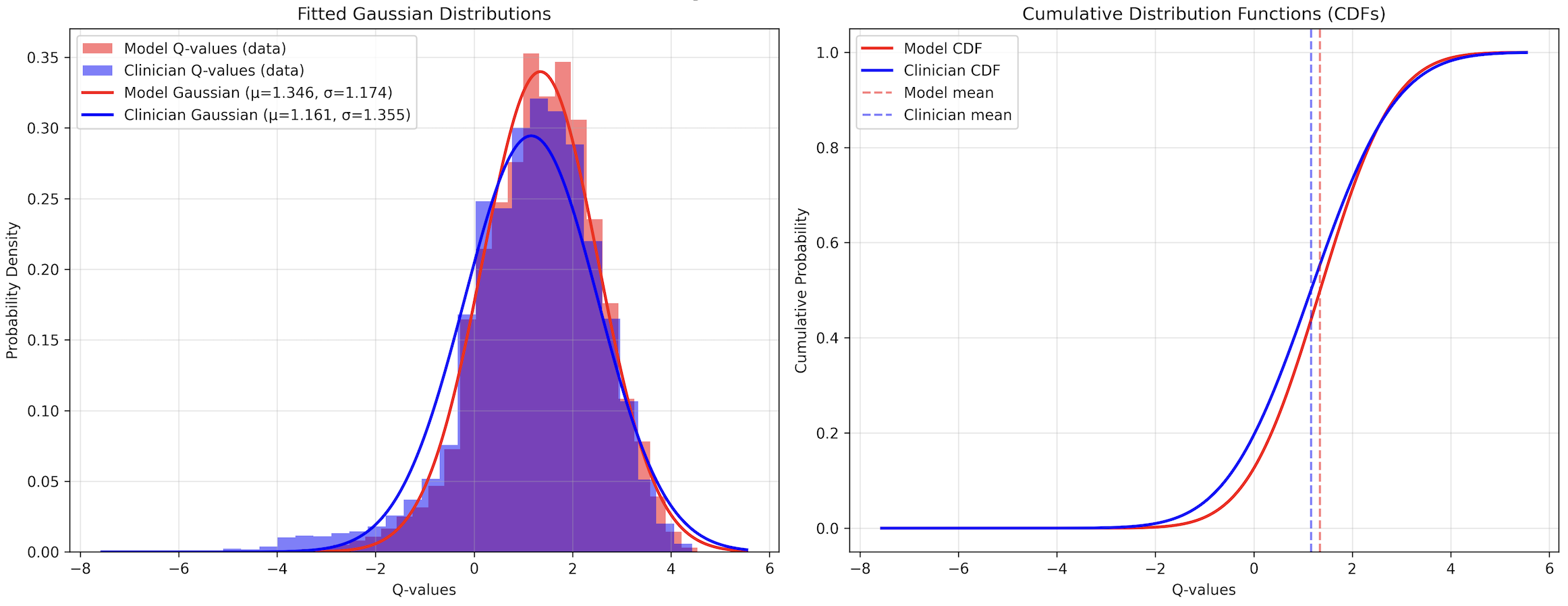}
\caption*{Dual Mixed Model: distribution of Q-values from model (red) and clinician (blue).}
\label{fig:q_histogram_dual}
\end{figure}

\begin{figure}[h]
\caption{FQE comparison between the learned patient model and clinician baseline (Dual BD).}
\centering
\includegraphics[width=0.70\textwidth]{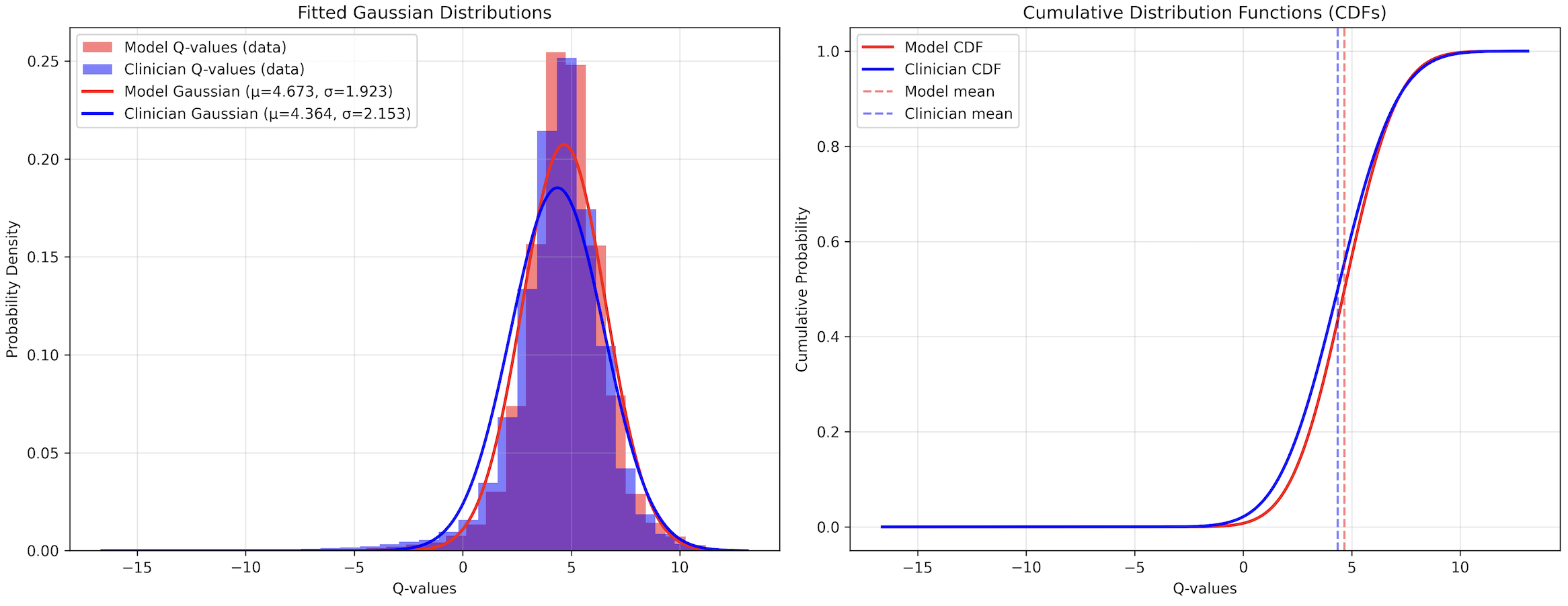}
\caption*{Block Discrete Model (10 bins): distribution of Q-values from model (red) and clinician (blue).}
\label{fig:q_histogram_block}
\end{figure}

\begin{figure}[h]
\caption{FQE comparison between the learned patient model and clinician baseline (Dual Stepwise).}
\centering
\includegraphics[width=0.70\textwidth]{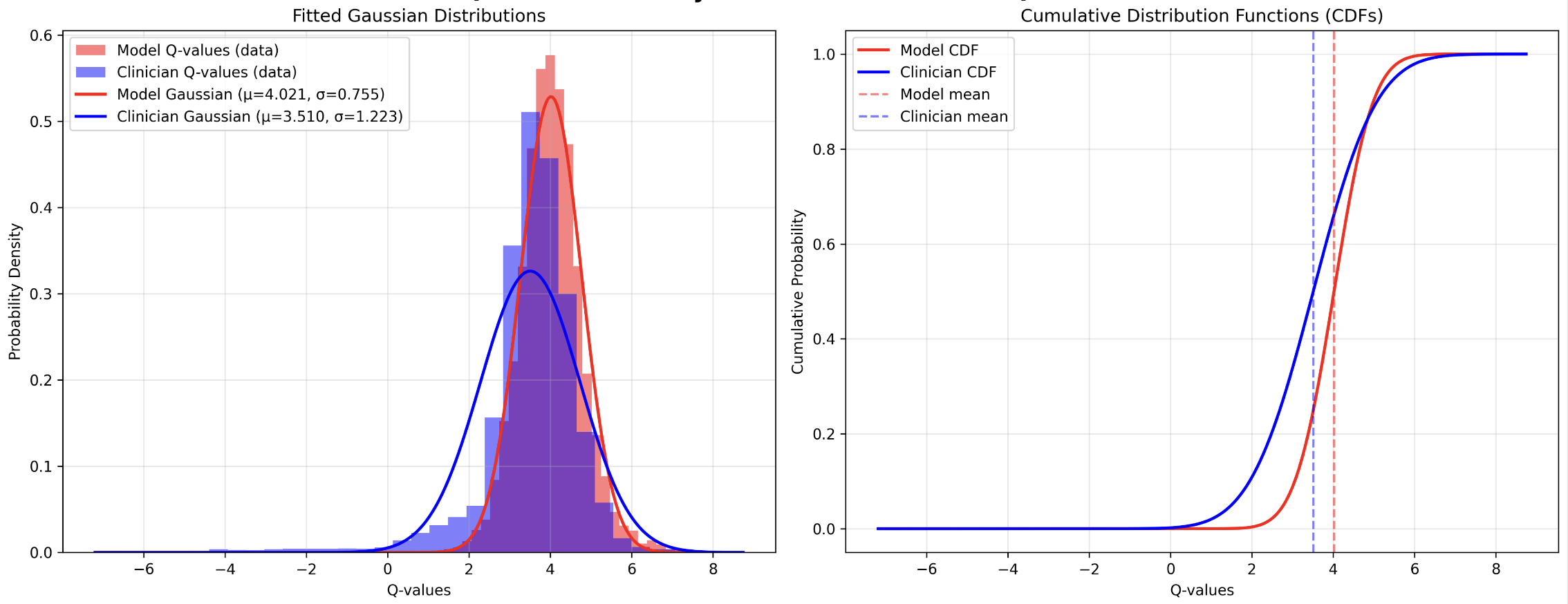}
\caption*{Dual Stepwise Model (max\_step=0.2): distribution of Q-values from model (red) and clinician (blue).}
\label{fig:q_cdfs}
\end{figure}

\begin{figure}[h]
\caption{FQE comparison between the learned patient model and clinician baseline (LSTM BD).}
\centering
\includegraphics[width=0.70\textwidth]{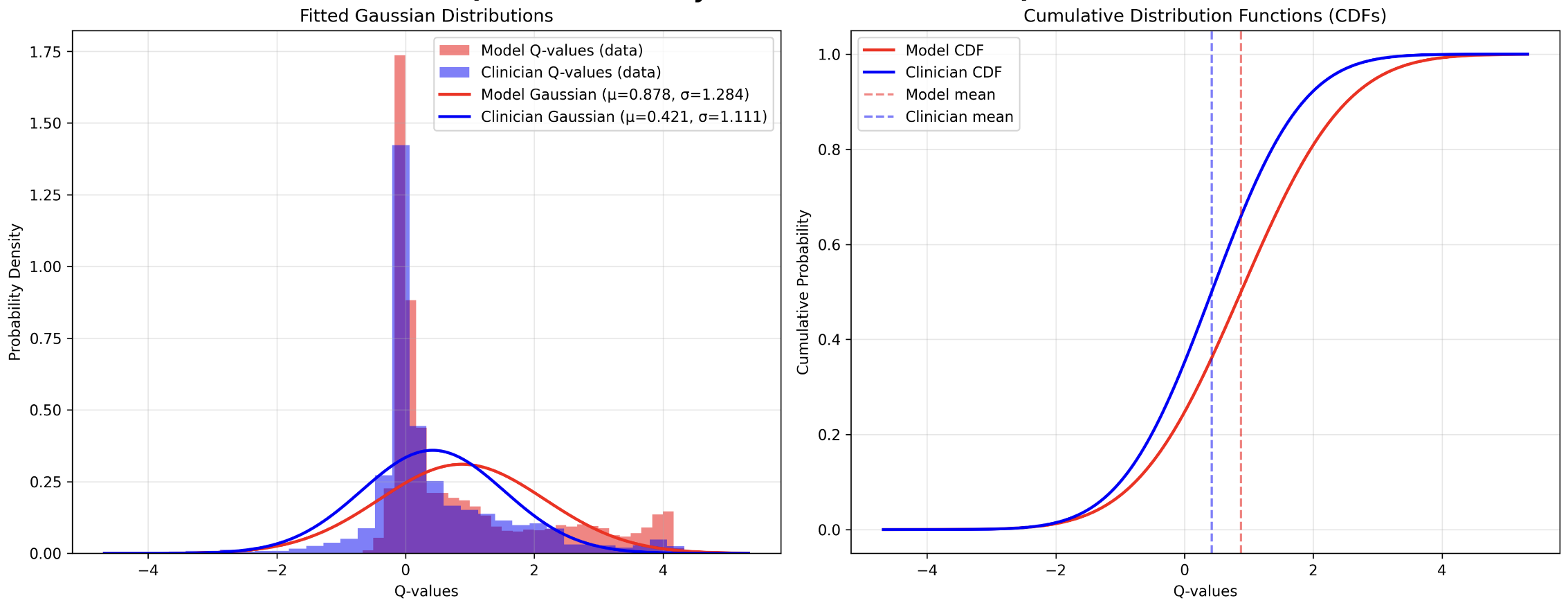}
\caption*{LSTM BD (10 bins): distribution of Q-values from model (red) and clinician (blue).}
\label{fig:q_improvement}
\end{figure}

\FloatBarrier

\section{Comprehensive Reward Function}
\label{apdx:crf}
We directly follow the OVISS~\cite{kalimouttou2025optimal} study to implement a full and comprehensive reward function with adjusted rewards. The reward 
measures patient progress and response to vasopressor administration in a quantifiable manner. The reward definition involves calculating an "adjusted reward" for each patient at different time points,
which is a composite measure reflecting both survival and physiological benefits.
\begin{itemize}
    \item \textbf{Survival Reward.} Each patient receives a base reward of $+1$ at every time step, reflecting the fundamental importance of maintaining survival.
    
    \item \textbf{Death Penalty.} If in-hospital mortality occurs, a penalty of $-20$ is applied at the terminal state of the trajectory. This large negative reward emphasizes the critical nature of mortality when evaluating treatment effectiveness.
    
    \item \textbf{Clinical Improvement Rewards.} Additional positive rewards are assigned when key physiological markers improve within prespecified future windows:
    \begin{itemize}
        \item \textbf{Lactate Decrease (+1):} A decrease in lactate levels within the next six hours, indicating improved metabolic status.
        \item \textbf{Mean Blood Pressure Increase (+1):} MBP rising to $\geq 65$\,mmHg within the next four hours, for cases where the initial MBP is below this threshold, reflecting improved hemodynamic stability.
        \item \textbf{SOFA Score Decrease (+3):} A reduction in SOFA score within the next six hours, highlighting improved multi-organ function.
        \item \textbf{Reduced Norepinephrine Usage (+1):} A decrease in norepinephrine dosage within the next four hours, suggesting better cardiovascular stability.
    \end{itemize}
\end{itemize}

\section{Detailed Experiment Results}
\label{apdx_sec:detailed_xp_results}

We present detailed comparisons between RL models and baselines.

\begin{table}[!ht]
\centering
\caption{Comprehensive comparison of Q-learning models including Binary, Dual Mixed, and Dual Block Discrete (with 3, 5, 10 bins) variants across different conservatism levels ($\alpha$).}
\label{tab:cql_comparison}
\begin{tabular}{llcccccc}
\toprule
Model & Config & $\alpha$ & $vp_1$ (\%) & Q/step & $\Delta$\textbf{Q/step} & $vp_1$ C. (\%) & $vp_2$ C. (\%) \\
\midrule
Clinician & -- & -- & 38.8 & 0.000 & 0.000 & 100.0 & -- \\
\midrule
Binary $vp_1$ & -- & 0.000 & 77.3 & 0.792 & 0.023 & 51.6 & -- \\
Dual Mixed & -- & 0.000 & 87.7 & 1.367 & 0.177 & 41.2 & -- \\
Dual BD & 3 bins & 0.000 & 71.4 & 1.734 & 0.123 & 55.2 & 57.1 \\
Dual BD & 5 bins & 0.000 & 96.3 & 2.383 & 0.265 & 39.7 & 26.4 \\
Dual BD & 10 bins & 0.000 & 92.3 & 4.673 & \textbf{0.309} & 42.4 & 13.3 \\
\midrule
Binary $vp_1$ & -- & 0.001 & 79.8 & 0.869 & 0.028 & 44.5 & -- \\
Dual Mixed & -- & 0.001 & 69.4 & 1.302 & 0.184 & 56.6 & -- \\
Dual BD & 3 bins & 0.001 & 82.2 & 1.614 & 0.110 & 46.9 & 55.5 \\
Dual BD & 5 bins & 0.001 & 93.6 & 2.281 & 0.217 & 41.3 & 27.2 \\
Dual BD & 10 bins & 0.001 & 96.6 & 4.445 & 0.292 & 39.1 & 14.1 \\
\midrule
Binary $vp_1$ & -- & 0.010 & 66.2 & 0.774 & 0.024 & 58.7 & -- \\
Dual Mixed & -- & 0.010 & 77.8 & 1.345 & 0.185 & 52.5 & -- \\
Dual BD & 3 bins & 0.010 & 76.8 & 1.576 & 0.127 & 53.1 & 56.3 \\
Dual BD & 5 bins & 0.010 & 84.9 & 2.091 & 0.119 & 43.9 & 40.3 \\
Dual BD & 10 bins & 0.010 & 92.8 & 4.072 & 0.245 & 41.4 & 16.6 \\
\bottomrule
\end{tabular}
\end{table}

\begin{table}[!ht]
\centering
\caption{Comparison of Binary $vp_1$ and Dual Stepwise  models with vasopressor persistence policy at different maximum step sizes.}
\label{tab:stepwise_comparison}
\begin{tabular}{lcccccc}
\toprule
Model & $\alpha$ & $vp_1$ (\%) & Q/step & $\Delta$Q/step & $vp_1$ C. (\%) & $vp_2$ C. (\%) \\
\midrule
Clinician & -- & 38.8 & 0.000 & 0.000 & 100.0 & -- \\
Binary $vp_1$ & 0.001 & 79.8 & 0.869 & 0.028 & 44.5 & -- \\
Dual Mixed & 0.010 & 77.8 & 1.345 & 0.185 & 52.5 & -- \\
\midrule
\multicolumn{7}{l}{max\_step = 0.1 (mcg/kg/min)} \\
\midrule
Dual Stepwise & 0.000000 & 34.6 & 0.046 & 0.103 & 61.1 & 17.9 \\
Dual Stepwise & 0.000100 & 12.4 & 0.094 & 0.136 & 65.1 & 17.7 \\
Dual Stepwise & 0.001000 & 82.9 & 0.252 & 0.128 & 47.4 & 24.5 \\
Dual Stepwise & 0.010000 & 45.9 & -0.337 & 0.114 & 60.7 & 22.3 \\
\midrule
\multicolumn{7}{l}{max\_step = 0.2 (mcg/kg/min)} \\
\midrule
Dual Stepwise & 0.000000 & 97.3 & 4.021 & \textbf{0.511} & 38.1 & 11.7 \\
Dual Stepwise & 0.000100 & 31.3 & 0.134 & 0.143 & 70.8 & 16.4 \\
Dual Stepwise & 0.001000 & 62.0 & 0.177 & 0.130 & 61.2 & 24.1 \\
Dual Stepwise & 0.010000 & 38.7 & -0.184 & 0.254 & 64.6 & 13.5 \\
\bottomrule
\end{tabular}
\end{table}

\section{Model Implementation and Training}
\label{apdx_sec:model_imp_training}
\subsection{Double Q-learning}
We use double-Q learning for all RL models in this paper. For our simplified version of the double-Q learning for stability, we adopt two Q-networks, and when Q-value is needed, the minimum scalar output across the two network is taken. Similarly, we adopt two target networks and take the minimum across them when computing the temporal difference (TD) error. 

\subsection{Binary vp1 Model}

The Binary $vp_1$ model represents the simplest action space, treating vasopressin administration as a binary decision (on/off). The $vp_2$ levels are used as part of the state. Training employs standard Q-learning TD loss with conservative logsumexp regularization over both action choices to prevent overestimation of out-of-distribution actions. The model processes states through fully connected networks with ReLU activations, using double Q-learning with target networks for stability. 

\subsection{Dual Mixed Model}

The Dual Mixed model handles two action dimensions: $vp_1$
(binary, 0 or 1) and $vp_2$ (continuous, 0-0.5 mcg/kg/min). This
continuous action space allows for fine-grained dosing control. For action selection of vp2, the model samples $50$ candidate actions per state during both training and inference, evaluating Q-values for each to select optimal doses. The conservative Q-learning penalty is computed through importance sampling over randomly sampled actions from the continuous space. 

\subsection{Dual Block Discrete Model}

The Dual Block Discrete model discretizes the continuous $vp_2$ action space into configurable bins while maintaining $vp_1$ as binary, creating a factored discrete action space with $2 \times N$ (combined with the
binary $vp_1$ action) total actions where $N$ is the number of $vp_2$
bins. The discretization uses uniform binning from $0$ to $0.5$ (mcg/kg/min)
with bin centers used for continuous dose reconstruction. Q-values are
computed for all discrete action combinations using one-hot encoding,
enabling efficient batch evaluation. The conservative Q-learning penalty applies logsumexp over all valid discrete actions to maintain conservative value estimates.

\subsection{Dual Stepwise Model}

The Dual Stepwise model implements an incremental dosing approach
where $vp_2$ changes are limited to fixed steps (e.g., single step $\pm$0.05,
max step$\pm$0.1 mcg/kg/min) from the current dose. This creates a
context-dependent action space that respects clinical constraints on
dose adjustments. The model augments state representations with
one-hot encoded current $vp_2$ dose bins and maintains valid action masks
to prevent out-of-bounds doses. The action space consists of $vp_1$
(binary) $\times$ $vp_2$ changes, with conservative Q-learning penalties computed only over valid actions given the current dose context.

\subsection{LSTM Block Discrete Model}

The LSTM Block Discrete model extends the block discrete approach with
sequential modeling capabilities. It processes patient trajectories
through LSTM layers to capture temporal dependencies in treatment
responses. The model uses sequence lengths of $5$ timesteps with $2$-step
burn-in periods for hidden state initialization. Training employs a
specialized replay buffer that maintains overlapping sequences with
mortality-weighted sampling priorities. The architecture combines LSTM
encoders ($2$ layers, $32$ hidden units) with Q-networks that process
concatenated LSTM outputs and state features. Actions are discretized
into $10$ norepinephrine dosing levels, with conservative Q-learning penalties computed over the discrete action space. 

\section{Model Architectures}
The model architectures for general Q-learning networks is shown in Table~\ref{model_arc:q_networks}. The model architectures for LSTM Block Discrete networks is shown in Table~\ref{model_arc:lstm_networks}.
\label{apdx_sec:model_arch}
\begin{table}[ht!]
\centering
\caption{Q Network Architecture (Binary/Mixed/Block/Stepwise)}
\label{model_arc:q_networks}
\begin{tabular}{ll}
\toprule
\textbf{Component} & \textbf{Architecture Details} \\
\midrule
\textbf{Input Layer} & State (\texttt{state\_dim}) + Action (\texttt{action\_dim}) \\
\midrule
\textbf{Hidden Layers} & [State, Action] $\xrightarrow{\text{FC (ReLU)}}$ 128 $\xrightarrow{\text{FC (ReLU)}}$ 128 $\xrightarrow{\text{FC (ReLU)}}$ 64 \\
\midrule
\textbf{Output Layer} & 64 $\xrightarrow{\text{FC}}$ 1 (Q-value) \\
\midrule
\textbf{Initialization} & Xavier Uniform \\
\midrule
\textbf{Function} & $Q(s,a) \rightarrow \mathbb{R}$ \\
\midrule
\textbf{Networks} & Dual Q-networks (Q1, Q2) with dual target networks \\
\bottomrule
\end{tabular}
\end{table}

\begin{table}[!ht]
\centering
\caption{LSTM Q Network Architecture (LSTM Block Discrete CQL)}
\label{model_arc:lstm_networks}
\begin{tabular}{ll}
\toprule
\textbf{Component} & \textbf{Architecture Details} \\
\midrule
\textbf{Input} & State (\texttt{state\_dim}) \\
\midrule
\textbf{Feature Extractor} & [State] $\xrightarrow{\text{FC (ReLU)}}$ 32 $\xrightarrow{\text{Dropout(0.1)}}$ \\
                          & $\xrightarrow{\text{FC (ReLU)}}$ 32 $\xrightarrow{\text{Dropout(0.1)}}$ \\
\midrule
\textbf{LSTM Layers} & 32 $\xrightarrow{\text{LSTM}}$ \texttt{2 layers} $\times$ 32 units \\
                     & (with dropout=0.1) \\
\midrule
\textbf{Output Layer (Q-head)} & 32 $\xrightarrow{\text{FC}}$ \texttt{num\_actions} (Q-values) \\
\midrule
\textbf{Initialization} & Xavier Uniform \\
\midrule
\textbf{Function} & $Q(s,a) \rightarrow \mathbb{R}^{\text{num\_actions}}$ \\
\midrule
\textbf{Networks} & Dual Q-networks with LSTM, dual target networks \\
\bottomrule
\end{tabular}
\end{table}

\section{Hyper-parameter Configurations}

\label{apdx_sec:hyperparameters}

The shared hyper-parameters are shown in Table~\ref{hp:shared}. The hyper-parameters specific to different action space models are shown in Table~\ref{hp:model_specific}. 


\begin{table}[!ht]
\centering
\caption{Common Hyperparameters Across All Models}
\label{hp:shared}
\begin{tabular}{lll}
\toprule
\textbf{Hyperparameter} & \textbf{Value} & \textbf{Description} \\
\midrule
Learning Rate (lr)     & $10^{-3}$ & Adam optimizer learning rate \\
Batch Size             & 128       & Training batch size \\
Gamma ($\gamma$)       & 0.95      & Discount factor for future rewards \\
Tau ($\tau$)           & 0.8       & Soft target network update rate \\
Gradient Clipping      & 1.0       & Maximum gradient norm \\
Epochs                 & 100       & Number of training epochs \\
Validation Batches     & 10        & Batches used for validation \\
Random Seed            & 42        & Ensures reproducibility \\
\bottomrule
\end{tabular}
\end{table}

\begin{table}[!ht]
\centering
\captionsetup{type=table}
\captionof{table}{Model Configurations: Action Spaces and Hyperparameters}
\label{hp:model_specific}
\begin{tabular}{@{}ll@{\hspace{0.5cm}}ll@{}}
\toprule
\multicolumn{2}{c}{\textbf{Action Space}} & \multicolumn{2}{c}{\textbf{Key Parameters}} \\
\midrule
\textbf{Model} & \textbf{Configuration} & \textbf{Model} & \textbf{Hyperparameters} \\
\midrule
Binary & 1D binary (0/1) & Binary & --- \\
Dual Mixed & 2D $vp_1$ binary, $vp_2$ continuous & Dual Mixed & \texttt{num\_samples}: 50 \\
           & $vp_1$: \{0,1\}, $vp_2$: [0,0.5] & & (continuous $vp_2$ selection) \\
Block Discrete & 2$\times$N discrete & Block Discrete & --- \\
               & $vp_1$: binary, $vp_2$: N bins & & \\
Stepwise & 2$\times$M discrete & Stepwise & \texttt{max\_step}: \{0.1, 0.2\} \\
         & $vp_1$: binary, $vp_2$: M steps & & \texttt{step\_size}: 0.05 \\
LSTM Block & N discrete ($vp_2$ only) & LSTM Block & \texttt{seq\_len}: 5, \texttt{burn\_in}: 2 \\
\bottomrule
\end{tabular}
\end{table}

\end{document}